\documentclass[sigconf]{aamas}  
\usepackage{booktabs}
\usepackage{array}
\usepackage{gensymb}
\usepackage[ruled,linesnumbered]{algorithm2e}
\SetKwInOut{Parameter}{Parameter(s)}

\usepackage{color, colortbl}
\definecolor{Gray}{gray}{0.9}

\usepackage{flushend}
\setcopyright{ifaamas}  
\acmDOI{doi}  
\acmISBN{}  
\acmConference[ALA'20]{Proc.\@ of the 12th Adaptive and Learning Agents (ALA 2020)}{May 2020}{Auckland, New Zealand}  
\acmYear{2020}  
\copyrightyear{2020}  
\acmPrice{}  

\renewcommand\footnotetextcopyrightpermission[1]{} 
\begin{document}
\title{Discrete-to-Deep Supervised Policy Learning}  
\subtitle{An effective training method for neural reinforcement learning}
\addtolength{\parskip}{-0.15mm}



%
\author{Budi Kurniawan}
\orcid{0000-0001-9016-8912}
\affiliation{
  \institution{Federation University}
  \city{Mount Helen} 
  \state{VIC} 
  \postcode{3350}
  \country{Australia}
}
\email{budikurniawan@students.federation.edu.au}

\author{Peter Vamplew}
\orcid{0000-0002-8687-4424}
\affiliation{
  \institution{Federation University}
  \city{Mount Helen} 
  \state{VIC} 
  \postcode{3350}
  \country{Australia}
}
\email{p.vamplew@federation.edu.au}

\author{Michael Papasimeon}
\orcid{0000-0003-1184-8376}
\affiliation{
  \institution{Defence Science and Technology Group}
  \city{Fishermans Bend}
  \state{VIC}
  \postcode{3207}
  \country{Australia}
}
\email{michael.papasimeon@dst.defence.gov.au}

\author{Richard Dazeley}
\orcid{0000-0002-6199-9685}
\affiliation{
  \institution{Deakin University}
  \city{Geelong}
  \state{VIC}
  \postcode{3220}
  \country{Australia}
}
\email{richard.dazeley@deakin.edu.au}

\author{Cameron Foale}
\orcid{0000-0003-2537-0326}
\affiliation{
  \institution{Federation University}
  \city{Mount Helen} 
  \state{VIC} 
  \postcode{3350}
  \country{Australia}
}
\email{c.foale@federation.edu.au}

\begin{abstract}  
Neural networks are effective function approximators, but hard to train in the reinforcement learning (RL) context mainly because samples are correlated. For years, scholars have got around this by employing experience replay or an asynchronous parallel-agent system. This paper proposes Discrete-to-Deep Supervised Policy Learning (D2D-SPL) for training neural networks in RL. D2D-SPL discretises the continuous state space into discrete states and uses actor-critic to learn a policy. It then selects from each discrete state an input value and the action with the highest numerical preference as an input/target pair. Finally it uses input/target pairs from all discrete states to train a classifier. D2D-SPL uses a single agent, needs no experience replay and learns much faster than state-of-the-art methods. We test our method with two RL environments, the Cartpole and an aircraft manoeuvring simulator.
\end{abstract}

\maketitle

\section{Introduction}
Table-based reinforcement learning (RL) techniques are easy to use but suffer from the classic curse of dimensionality, which means they are powerless to fix problems with high dimensional spaces. For this type of problem, we need RL methods that use function approximation, a technique commonly used in supervised learning (SL) \cite{Sutton2018}. The neural network is a popular function approximator in SL and many scholars have tried to use it in RL. The first success story of combining RL with a neural network is that of TD-Gammon, a computer program that plays backgammon \cite{Tesauro1995}. Despite this initial success, however, subsequent applications of neural network-based RL, or neural RL, to solve large-scale RL problems failed, which led to the abandonment of neural RL. Some consider TD-Gammon a one-time exception and believe the success of TD-Gammon was not due to the RL algorithm used, but because of the setup of co-evolutionary self-play biased by the dynamics of backgammon \cite{Pollack1997}.

The inability of neural networks to act as good function approximators in many RL studies is mainly attributed to the fact that samples generated in RL learning are correlated and not independently and identically distributed. Additionally, the neural network often forgets what it has learned, forcing it to re-learn when it sees a sample that has previously been presented to it, something known as the relearning problem \cite{Lin1993}.

There are two ways to overcome these shortcomings, by using experience replay \cite{Lin1993} or by employing multiple agents working asynchronously to update a centralised network \cite{Mnih2016}. Both have proven effective but have weaknesses as well. Experience replay with a large buffer can hurt performance and both a small and a large replay buffer can hurt the learning process \cite{Zhang2017}. Asynchronous parallel agents are sample inefficient and require a complex application architecture \cite{Wang2016}.

In this paper we propose discrete-to-deep supervised policy learning (D2D-SPL), a variant of supervised policy learning, to successfully train a neural network as a function approximator to solve high-dimensional RL problems. D2D-SPL is simple and based on the standard actor-critic with eligibility traces algorithm \cite{Barto1983}. Our tests show that our method learns much faster than techniques that are based on experience replay and parallel agents.

\section{Related Work}
The first successful large-scale neural RL application is TD-Gammon \cite{Tesauro1995}, a computer program that plays backgammon. TD-Gammon learned a policy by playing against itself. To get to the level where it could beat another computer-based backgammon player, it self-played 300,000 games. To make it even more skilful to the point it could defeat a grandmaster, it self-played 1.5 million games. After this initial success, however, researchers attempting to exploit neural networks to solve large-scale RL problems were faced with repeated failures, as reported in \cite{Boyan1995}. This resulted in a move away from neural RL, before several techniques were proposed that led to future successes. Those techniques included methods based on experience replay \cite{Lin1993}, such as Neural Fitted Q Iteration (NFQ) \cite{Riedmiller2005} and DQN \cite{Mnih2013}, and methods based on asynchronous parallel agents such as asynchronous advantage actor-critic (A3C) \cite{Mnih2016} and actor-critic with experience replay (ACER) \cite{Wang2016}.

Experience replay is an interesting invention because it inspires successful large-scale neural RL cases that came many years later. An experience is a tuple ($s$, $a$, $r$, $s'$), where $s$ is the state the agent is in while taking action $a$, resulting in reward $r$ and a new state $s'$. Experience replay is a method whereby a collection of experiences are presented repeatedly to the learning algorithm as if the learning agent experienced it again and again. Two benefits of experience replay are the expedition of the process of credit assignment and the opportunity for the agent to refresh what it has learned \cite{Lin1993}.

NFQ is based on experience replay and uses a multi-layer perceptron (MLP). The principle idea of NFQ is to collect all experience tuples within an episode and present the collection at the end of the episode to train the MLP rather than after each timestep. Targets are generated using a cost function. This is repeated for a preset number of episodes and the experience collection at each episode includes all the collections from prior episodes \cite{Riedmiller2005}.

One notable success of experience replay is its use in deep Q-network (DQN) for playing ATARI games, where incoming samples are stored in a buffer and a subset of these samples are randomly selected from the buffer at every timestep to train the neural network \cite{Mnih2013}.

The same team extends DQN by introducing a second network called the target network to better de-correlate samples. At the beginning of a learning episode, the primary network is cloned and the clone is used to produce targets for the primary network. Every $n$ timesteps the weights of the target network are updated using those of the primary network. This technique is known as DQN with a target network \cite{Mnih2015}. Its success can be attributed to two innovations: 1. off-policy training with samples from a replay buffer to minimise correlations between samples; 2. The use of a target Q network to give consistent targets during temporal difference backups. 

Other derivatives of DQN include a parallel implementation called Gorila \cite{Nair2015}, double DQN \cite{vanhasselt2016}, prioritised experience replay \cite{Schaul2015}, dueling D-DQN \cite{Wang2015}, distributed prioritised experience replay \cite{Horgan2018} and deep deterministic policy gradient (DDPG) \cite{Lillicrap2015}, which covers continuous action spaces.
Rainbow, another variant of DQN, combines features in other DQN variants and proves able to improve performance in some ATARI games \cite{Hessel2018}. In addition, ACER \cite{Wang2016} improves A3C's sample efficiency by using multiple agents and experience replay.

The main difference between our work and the DQN family of techniques is that our method does not use experience replay. Our method works with discrete action spaces, unlike DDPG that is suitable for continuous action spaces. Unlike NFQ, our method only trains the neural network once with data obtained when the actor-critic policy has somewhat stabilised.

Using data generated from a RL policy to train a neural network has also been used in studies called guided policy search \cite{Levine2015} and supervised policy learning \cite{Chebotar2016}. Both studies are different from ours because they used model-based RL, whereas our RL algorithm, actor-critic, is model-free. Additionally, they used data from multiple policies to train a classifier whereas our technique only requires one policy. Furthermore, in \cite{Chebotar2016} an optimisation strategy is needed before data can be fed to the neural network. Our technique does not require optimisation between the RL part and the classifier.

A technique called supervised actor-critic \cite{Rosenstein2002} and its variant \cite{Wang2018} also combine RL and SL. In this architecture, the actor receives a signal from the critic as well as from a neural network. By contrast, D2D-SPL is a two-step process that first uses standard actor-critic and then selects the data generated from the first step to train a classifier.

Other scholars have used classifiers in RL, including \cite{Lagoudakis2003} who use a support vector machine in their RL algorithm. Their use of a classifier is restricted to the inner loop of their RL algorithm, unlike D2D-SPL, which uses a classifier in action selection. 
\section{Discrete-to-Deep Supervised Policy Learning (D2D-SPL)}
Combining RL and SL, D2D-SPL is suitable for solving continuous-state RL problems with discrete actions. It uses off-policy data from an actor-critic policy to train a neural network that thereafter can be used as a controller for an RL problem. D2D-SPL works in two phases, a RL phase and a SL phase. First, it discretises the continuous state space and learns a policy using actor-critic with eligibility traces. This policy, which is based on a coarse discretisation, should be able to be learned more quickly than a policy based on the full, continuous state-space. Second, it uses data generated during reinforcement learning to train a classifier. Not all samples are used. The method selects from each discrete state an input value and the action with the highest preference as an input/target pair. The classifier learns when all input/target pairs are presented to it at the same time, thus eliminating the need for online learning. 

To use D2D-SPL, we start by discretising the state space into discrete states. The number of discrete states varies depending on the complexity of the problem. For example, our Cartpole solution can achieve its targets with 162 discrete states. By contrast, the aircraft manoeuvring simulation problem that we use to test our algorithm requires 14,000 states to learn a good policy. As will be seen shortly, the number of discrete states is also the maximum number of samples for the second-stage supervised learning. The number of continuous state variables is the same as the number of input nodes to the classifier.

Once discrete states are identified, we start learning by using  actor-critic with eligibility traces. In every episode we group state variables by discrete state and store the total reward of the episode. Once learning is finished, we select the top 5\% episodes having the highest total rewards and average the values of each state variable in each discrete state. We end up with $n$ samples as inputs to the neural network, where $n$ <= the number of discrete states. $n$ can be lower than the number of discrete states because we filter out discrete states that were never visited during reinforcement learning.

Algorithm 1 shows the reinforcement phase of our method. It is basically the actor-critic with eligibility traces algorithm \cite{Barto1983} with a buffer for storing tuples of state variable values and the number of times a discrete state has been visited. We then use the buffer and the resulting policy as inputs to the supervised\_phase function in Algorithm 2, which shows how samples are selected and prepared for training the classifier.

\begin{algorithm}
\SetAlgoLined
\KwIn{a differentiable policy parameterisation $\pi (a|s, \boldsymbol{\theta})$}
\KwIn{a differentiable state-value function parameterisation $\hat{v} (s, \boldsymbol{\textrm{w}})$}
\Parameter{number of discrete states $N_{ds}>$ 0, number of actions $N_a>$ 0, number of state variables $N_{sv}>$ 0}
\Parameter{step sizes $\alpha^{\boldsymbol{\theta}}$ $>$ 0, $\alpha^{\boldsymbol{\textrm{w}}}$ $>$ 0}
\Parameter{trace decay rates $\lambda^{\boldsymbol{\theta}} \in$ [0,1], $\lambda^{\boldsymbol{\textrm{w}}} \in$ [0,1]}
Initialise policy parameter $\boldsymbol{\theta} \in \mathbb{R}^{N_{ds}xN_a}$ and state-value weights $\boldsymbol{\textrm{w}} \in \mathbb{R}^{N_{ds}}$ (e.g, to 0)\;
Initialise empty buffer $B$\;
 \For{each episode}{
    Initialise $S$ (first state of episode)\;
    Initialise $M_{sv}\in \mathbb{R}^{N_{ds}xN_{sv}}$ for storing aggregate values of state variables\;
    Initialise $M_{sc}\in \mathbb{R}^{N_{ds}}$ for keeping track of the number of times a value has been added to $M_{sv}$\;
    $\textrm{z}^{\boldsymbol{\theta}} \leftarrow \boldsymbol{0}$ ($N_{ds}$x$N_a$-component eligibility trace matrix)\;
    $\textrm{z}^{\boldsymbol{\textrm{w}}} \leftarrow \boldsymbol{0}$ ($N_{ds}$-component eligibility trace vector)\;
    $I \leftarrow$ 1 \;

    $R_{total} \leftarrow$ 0\;
    \For{each step of episode or until $S$ is terminal}{
      $S_{disc}\leftarrow$ discretise($S$)\;
      $A \sim \pi(.|S_{disc}, \boldsymbol{\theta})$\;
      Take action $A$, observe $S'$, $R$\;
      $M_{sv}[S_{disc}] \leftarrow M_{sv}[S_{disc}]$ + $S$\;
      $M_{sc}[S_{disc}] \leftarrow M_{sc}[S_{disc}]$ + 1\;
      $R_{total} \leftarrow R_{total}$ + $R$\;
      $S'_{disc}\leftarrow$ discretise($S'$)\;
      $\delta \leftarrow R + \gamma \hat{v}(S'_{disc},\boldsymbol{\textrm{w}}) -  \hat{v}(S_{disc},\boldsymbol{\textrm{w}})$ \ \ \  (if $S'$ is terminal, $\hat{v}(S'_{disc},\boldsymbol{\textrm{w}}) \doteq$ 0)\;
      $\textrm{z}^{\boldsymbol{\textrm{w}}} \leftarrow \gamma \lambda^{\boldsymbol{\textrm{w}}} \textrm{z}^{\boldsymbol{\textrm{w}}} + \nabla \hat{v}(S_{disc},\boldsymbol{\textrm{w}})$\;
      $\textrm{z}^{\boldsymbol{\theta}} \leftarrow \gamma \lambda^{\boldsymbol{\theta}} \textrm{z}^{\boldsymbol{\theta}} + I \nabla \ln \pi (A|S_{disc}, \boldsymbol{\theta})$\;
      $\boldsymbol{\textrm{w}} \leftarrow \boldsymbol{\textrm{w}} + \alpha^{\boldsymbol{\textrm{w}}} \delta \boldsymbol{\textrm{z}^{\boldsymbol{\textrm{w}}}}$\;
      $\boldsymbol{\theta} \leftarrow \boldsymbol{\theta} + \alpha^{\boldsymbol{\theta}} \delta \boldsymbol{\textrm{z}}^{\boldsymbol{\theta}}$\;
      $I \leftarrow \gamma I$\;
      $S \leftarrow S'$\;
    }
    Add ($R_{total}$, $M_{SV}$, $M_{SC}$) to $B$\;
 }
Call $supervised\_phase$($B$, $\boldsymbol{\theta}$)
 \caption{D2D-SPL reinforcement phase}
\end{algorithm}

\begin{algorithm}
\SetAlgoLined
\KwIn{buffer containing tuples of ($R_{total}$, $M_{sv}$, $M_{cv}$)}
\KwIn{policy parameter $\boldsymbol{\theta}$}
\Parameter{number of discrete states $N_{ds}>$ 0, number of state variables $N_{sv}>$ 0}
Sort $B$ by $R_{total}$ in descending order and delete the bottom 95$\%$ elements\;
Initialise $ConsM_{sv}\in \mathbb{R}^{N_{ds}xN_{sv}}$ for storing consolidated values of $M_{sv}$\;
Initialise $ConsM_{sc}\in \mathbb{R}^{N_{ds}}$ for storing consolidated values of $M_{sc}$\;
Initialise $T\in \mathbb{R}^{N_{ds}}$ for storing targets for the classifier\;
\For{each element $E$ in $B$} {
    $R_{total}$, $M_{SV}$, $M_{SC}\leftarrow E$\;
    \For{$i \leftarrow$ 0 To $N_{ds}$ - 1} {
        $ConsM_{SV}[$i$]\leftarrow ConsM_{SV}$[$i$] + $M_{SV}$[$i$]\; 
        $ConsM_{SC}[$i$]\leftarrow ConsM_{SC}$[$i$] + $M_{SC}$[$i$]\; 
    }
}
\For{$i \leftarrow$ $N_{sd}$ - 1 To 0} {
    \eIf{$ConsM_{sc}$[$i$] == 0}{
        Delete element $i$ from $ConsM_{sv}$ and $T$\;
    }{
        $ConsM_{sv}$[$i$]$\leftarrow ConsM_{sv}$[$i$] $/$ $ConsM_{sc}$[$i$]\;
        $T$[$i$] $\leftarrow$ index of the largest value of $\boldsymbol{\theta}$[$i$]
    }
}
Use $ConsM_{SV}$ as inputs and $T$ as targets to train classifier\;
 \caption{D2D-SPL supervised phase}
\end{algorithm}

\begin{figure}[h!]
  \centering
  \includegraphics[width=6cm,keepaspectratio]{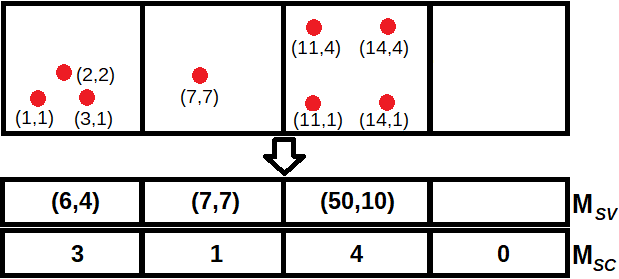}
  \caption{Data consolidation in every episode of D2D-SPL}
\end{figure}

Figure 1 shows how data is consolidated every episode. For simplicity it shows a case where there are only four discrete states (each represented by a box at the top diagram) and there are two state variables in each state. At every timestep, the values of the state variables of the visited state are recorded. Figure 1 shows the data collected after eight timesteps, in which Discrete State 1 has been visited three times with state variable tuples (1,1), (2,2) and (3,1), Discrete State 2 visited once with state variable tuple (7,7) and so on. Since we are only concerned with the average values of state variables in each discrete state, we can save memory by just keeping the totals of state variable values in every discrete state in $M_{SV}$ and the number of times that state is visited in $M_{SC}$. Along with the total reward for the episode, we insert the tuple ($R_{TOTAL}$, $M_{SV}$, $M_{SC}$) to buffer $B$. At the end of Algorithm 1, the number of tuples in buffer $B$ will be the same as the number of episodes run.

Buffer $B$ and the actor-critic policy are passed to function $supervised\_phase$ in Algorithm 2. The buffer contains the fore-mentioned tuples and the policy contains numerical preferences (one for each action) for each discrete state. The objective of this function is to produce an input/target pair for every discrete state, excluding states that were never visited during training. The function starts by selecting the top 5\% tuples with the highest total rewards in the buffer and consolidate the selected tuples into at most one tuple for each discrete state. Data is consolidated by summing the values of each state variable in a discrete state and divided them by the number of times the state was visited. For each discrete state, we look up the policy and select as a target the action with the highest numerical preference. We remove all states that are not present in the input set from the target set and pass the training set to the classifier.

\section{Experiments and Discussion}
We test our method with two RL environments, the classic Cartpole and an aircraft manoeuvring simulator. We choose Cartpole because it is a well-known problem and the aircraft manoeuvring simulator because it represents a large-scale problem that is not easy to solve using table-based methods.

In each of the two problems, the agent learns an actor-critic policy for $n$ episodes. We call the policy it learns Discrete $n$ and use it as the base for several other solutions. In the first solution we clone the policy and use the clone as the base policy to run another $n$ episodes of actor-critic learning and call the resulting policy Discrete 2$n$, because it is the result of learning for 2$n$ episodes. In the second solution, which we call D2D-SPL, we use the data generated from Discrete $n$ to train a classifier.

Separately, for comparison, we use DQN \cite{Mnih2013}, Double DQN (DDQN) \cite{vanhasselt2016} and A3C \cite{Mnih2016} to solve the same problems. A3C, despite it being four years old, is currently the state-of-the-art method. We use four parallel agents in all our A3C tests.

For Cartpole, we use eight systems to compare: Discrete 1,000, Discrete 2,000, D2D-SPL, DQN 1,000, DQN 2,000, DDQN 1,000, DDQN 2,000 and A3C 4,000. For Aircraft Manoeuvring we use Discrete 20,000, Discrete 40,000, D2D-SPL, DQN 20,000, DQN 40,000, DDQN 20,000, DDQN 40,000, A3C 20,000 and A3C 200,000. All experiments are run on an Intel i9-7900X (10 cores, 20 threads) machine with two Nvidia 1080 GTX cards. For all the methods involving neural networks, we use a relatively simple architecture containing a single hidden layer with twelve nodes in Cartpole and fifty nodes in Aircraft Manoeuvring.

\subsection{Cartpole}
Cartpole is a pole-balancing control problem posed in \cite{Michie1968} for which Barto et. al. proposed a solution using actor-critic reinforcement learning \cite{Barto1983}. A solution aims to control the free-moving cart to which the pole is attached by exerting a force to the left or to the right of the cart to keep the pole standing. We use the OpenAI Gym Cartpole environment \cite{Brockman2016} that is based on Barto et. al.'s solution but is different from the original system in that the four state variables in Gym are randomised at the beginning of every episode, whereas in the original system the variables are always set to zero.

OpenAI Gym's Cartpole comes in two flavours, one that considers the problem solved if the pole remains standing for 200 consecutive timesteps and another whose target is 500 timesteps. We make the problem more complex by raising the target to 100,000.

Our code is published on https://github.com/budi-kurniawan/d2d-spl.

The agent gets a +1 reward for each timestep, including when the agent lands on a terminal state. Each episode is terminated when one of these two things occurs: When the pole falls (failure) or when the target is achieved (success). Therefore, the time spent in each episode is proportional to the number of total rewards. The more successful learning in an episode is, the longer the episode takes.

For each method we run ten trials that each involves training the agent once, and then testing the final policy or model on 100 different runs from different starting positions. We use the same random seeds for all methods, making sure the initial values for all methods are the same for each trial. The average rewards for all runs are shown in Table 1. Table 2 shows how many tests in each trial achieve the target reward of 100,000.
\begin{table*}
\centering
\begin{tabular}{|c|r|r|r|r|r|r|r|r|} 
\hline
Trial & Discrete & Discrete & \multicolumn{1}{|c|}{D2D} & \multicolumn{1}{|c|}{DQN} & \multicolumn{1}{|c|}{DQN} & \multicolumn{1}{|c|}{DDQN} & \multicolumn{1}{|c|}{DDQN} & \multicolumn{1}{|c|}{A3C}\\
      & \multicolumn{1}{|c|}{1,000} & \multicolumn{1}{|c|}{2,000} 
      & \multicolumn{1}{|c|}{SPL} & \multicolumn{1}{|c|}{1,000} & \multicolumn{1}{|c|}{2,000} & \multicolumn{1}{|c|}{1,000} & \multicolumn{1}{|c|}{2,000} & \multicolumn{1}{|c|}{4,000}\\
\hline
0 & 443.42 & 2395.40 & \textbf{41,205.76} & 9.31 & 46.57 & 426.51 & 13.78 & 17.83\\
1 & 253.01 & 443.85 & 636.60 & 70.61 & 1,476.75 & \textbf{69,049.44} & 163.32 & 24.47\\
2 & 378.59 & 495.04 & 96,004.04 & 9.48 & 116.52 & 83,064.61 &	\textbf{100,000.00} & 93.93\\
3 & 241.19 & 208.69 & 32.01 & 11.47 & 30.17 & 143.71 & \textbf{78,132.04} & 12.02\\
4 & 206.49 & 210.14 & 14,194.83 & 16.09 & 10.80 & \textbf{100,000.00} & 288.71 & 76.05\\
5 & 130.79 & 144.49 & \textbf{18,623.59} & 28.81 & 30.61 & 9.71 & 243.64 & 10.32\\
6 & 2,177.08 & \textbf{2,581.67} & 1,098.02 & 207.67 & 9.42 & 18.02 & 446.12 & 12.90\\
7 & 349.99 & 230.72 & \textbf{100,000.00} & 178.64 & 69.46 & 92.76 & 69.40 & 14.34\\
8 & 137.32 & 112.44 & 86.46 & 437.77 & 23.53 & \textbf{100,000.00} & 191.21 & 9.47\\
9 & 349.23 & 363.38 & 12.28 & \textbf{7,160.00} & 11.14 & 89.90 & 189.81 & 119.22\\
\hline
Mean & 466.71 & 718.58 & 27,189.36 & 812.99 & 182.50 & \textbf{35,289.47} & 17,973.80 & 39.06\\
\hline
Median & 301.12 & 297.05 & \textbf{7,646.43} & 49.71 & 30.39 & 285.11 & 217.43 & 16.09\\
\hline
\end{tabular}
\caption{Average rewards for all Cartpole solutions}
\vspace{-0.1cm}
\centering
\begin{tabular}{|c|r|r|r|r|r|r|r|r|} 
\hline
Trial & Discrete & Discrete & \multicolumn{1}{|c|}{D2D} & \multicolumn{1}{|c|}{DQN} & \multicolumn{1}{|c|}{DQN} & \multicolumn{1}{|c|}{DDQN} & \multicolumn{1}{|c|}{DDQN} & \multicolumn{1}{|c|}{A3C}\\
      & \multicolumn{1}{|c|}{1,000} & \multicolumn{1}{|c|}{2,000} 
      & \multicolumn{1}{|c|}{SPL} 
      & \multicolumn{1}{|c|}{1,000} & \multicolumn{1}{|c|}{2,000} 
      & \multicolumn{1}{|c|}{1,000} & \multicolumn{1}{|c|}{2,000} 
      & \multicolumn{1}{|c|}{4,000}\\
\hline
0 & 0 & 0 & \textbf{41} & 0 & 0 & 0 & 0 & 0\\
1 & 0 & 0 & 0 & 0 & 1 & \textbf{69} & 0 & 0\\
2 & 0 & 0 & 96 & 0 & 0 & 83 & \textbf{100} & 0\\
3 & 0 & 0 & 0 & 0 & 0 & 0 & \textbf{78} & 0\\
4 & 0 & 0 & 14 & 0 & 0 & \textbf{100} & 0 & 0\\
5 & 0 & 0 & \textbf{6} & 0 & 0 & 0 & 0 & 0\\
6 & 0 & 0 & \textbf{1} & 0 & 0 & 0 & 0 & 0\\
7 & 0 & 0 & \textbf{100} & 0 & 0 & 0 & 0 & 0\\
8 & 0 & 0 & 0 & 0 & 0 & \textbf{100} & 0 & 0\\
9 & 0 & 0 & 0 & \textbf{7} & 0 & 0 & 0 & 0\\
\hline
\end{tabular}
\caption{Successes in solving the Cartpole problem}
\end{table*}

Tables 1 and 2 show that D2D-SPL outperforms the Discrete 1,000 agent on which it is based, demonstrating that the neural network's ability to generalise from the discrete policy is beneficial. D2D-SPL is also more effective than all the other methods, except DDQN. A3C, being sample-inefficient and learned in only 4,000 episodes, performs much worse than the other solutions.

\subsection{Aircraft Manoeuvring Simulator}
\subsubsection{The Environment}
The second environment we test our method with is an air combat simulator called Ace Zero, which was developed by Defence Science and Technology Group, Australia and used in \cite{Ramirez2018}, \cite{Masek2018} and \cite{Lam2019}. We set the simulator for one-on-one fights in two-dimensional space, representing a continuous space sequential-decision problem much larger than Cartpole. In this domain we aim to develop an agent that can learn to execute aerial manoeuvres for autonomous aircraft. The goal is for our pilot agent to learn to pursue another autonomous aircraft, that is itself manoeuvring. The basic manoeuvre that we are exploring is know as a ``pure pursuit'' manoeuvre and can be considered a basic building block for more complex aerial manoeuvres such as formation flying or even within visual range air to air combat. We would like our pilot agent not only to learn how to manoeuvre, but also to adapt to the manoeuvres of a dynamic opponent.

For our agent, the environment offer an action space with five discrete actions: do nothing, turn left by 10\degree, turn right by 10\degree, increase speed by 10\% and decrease speed by 10\%. The opposing agent, however, is allowed to perform continuous change of speed and direction within its physical limit. Our desired goal state is for our learning pilot to manoeuvre their aircraft to be in a specific relative geometry configuration with respect to the aircraft being pursued. To describe this geometry, we use the standard measurements such as the range, the attack angle (AA) and the antenna train angle (ATA). Figure 2 shows this geometry. By convention, blue is used to depict the aircraft controlled by the subject agent and red to represent the opposing aircraft. The angles are shown from the point of view of the blue aircraft.

\begin{figure}[h!]
  \centering
  \includegraphics[width=5cm,keepaspectratio]{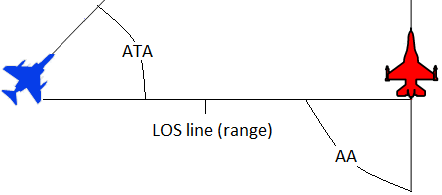}
  \caption{Relative aircraft geometry}
\end{figure}

The aircraft centres of mass are connected by the line of sight (LOS) line, which is also used to calculate the range between the two aircraft. The aspect angle (AA) is the angle between the LOS line and the tail of the red aircraft. It is an angular measure of how far the pursuer is of the pursued aircraft's tail. The antenna train angle (ATA) is the angle between the nose of the blue aircraft and the LOS line. It is an angular measure of how far the pursued aircraft is off the aircraft nose of the pursuer. The value of the AA and ATA is within 0\degree and 180\degree. By convention, angles to the right side of the aircraft are considered positive and angles to the left negative \cite{McGrew2010}.

The range, AA, ATA and the speed difference between the two aircraft are the four state variables making up each state.

The reward for each action taken is proportional to how favourable our agent's position is relative to the opposing agent, which can be measured quantitively using the McGrew score \cite{McGrew2010}. The score incorporates the range, AA and ATA. The McGrew score consists of two components, McGrew angular score ($A_M$) and McGrew range score ($R_M$).

\begin{equation}
S_M = A_M R_M
\end{equation}

The McGrew angular score is defined as follows.
\begin{equation}
A_M = \frac{1}{2}[(1 - \frac{AA}{180^\circ}) + (1 - \frac{ATA}{180^\circ})]
\end{equation}
Here, $AA$ and $ATA$ are in degrees and described in Figure 2. The maximum possible value for $A_M$ is 1, which is achieved when $AA$ = $ATA$ = 0.

The McGrew range score is defined as this.
\begin{equation}
R_M = exp[- \frac{|R-R_d|}{k \times 180^\circ}]
\end{equation}

where $R$ is the current range of the two aircraft and $R_d$ the desired range. $R_d$, the midpoint between the minimum gun range (500 feet = ~153m) and the maximum gun range (3,000 feet = ~914m), is about 380m \cite{Shaw1985}. The value of k, the hyper-parameter scaling factor, determines the width of the function peak around $R_d$. The larger the value of $k$, the bigger the spread. A small value of $k$ dictates that a high McGrew range score can only be achieved if the two aircraft are very close to the desired range. By default $k$ = 5.

The McGrew score ranges from 0.0 to 1.0 (inclusive). It approaches 1.0 when our agent is following the opponent within the desired range. By contrast, when it is being pursued by the opponent, the McGrew score is close to 0.

For all agents, we offset the McGrew score by -0.5 to make them learn faster as described in our previous study \cite{Kurniawan2019}.

\begin{equation}
Reward = S_M - 0.5
\end{equation}

\subsubsection{Test Results}
For the discretised actor-critic solutions, we discretise the state space into 14,000 discrete states. The range is split into fourteen regions, the AA into ten regions, the ATA into ten regions, and the speed difference into ten regions, resulting in 14,000 states. 
For training for all solutions, we start the opponent (Red aircraft) from position ($x_r$, $y_r$, $\psi_r$) where $x_r$ and $y_r$ are a coordinate in a Cartesian coordinate and $\psi_r$ the flying direction (heading) in degrees (relative to the X axis). Our aircraft (Blue) always starts from the origin with heading 0\degree and an initial speed of 125m/s, which means it starts by flying along the X axis. Red always starts from (1500$\pm\Delta$, 300$\pm\Delta$, 50\degree$\pm\Delta$), where $\Delta$ is a small positive random number, and flies in a straight line at a constant speed of 125m/s. The initial positions, headings and speeds of both aircraft are the same for all episodes. All episodes are terminated after 700 timesteps. 

We start by running the agent for 20,000 episodes, resulting in Discrete 20,000. This base policy is then cloned for the same agent to continue learning another 20,000 agents, resulting in Discrete 40,000. The base is also used for D2D-SPL. Separately, we use DQN, DDQN and A3C to produce DQN 20,000, DQN 40,000, DDQN 20,000, DDQN 40,000, A3C 20,000 and A3C 200,000. The choice for 200,000 episodes for the second A3C solution is due to the fact that A3C is data inefficient and need much more episodes to achieve comparable scores.

The policies from Discrete 20,000 and Discrete 40,000 as well as models from D2D-SPL, DQN, DDQN and A3C solutions are then used to test agents against an opponent that flies along paths that were not seen during training. Figures 3 to 6 show four test scenarios applied against the D2D-SPL models. The red paths represent the opponent's trajectories and the blue ones our agent's. 

\begin{figure}
\includegraphics[width=2in]{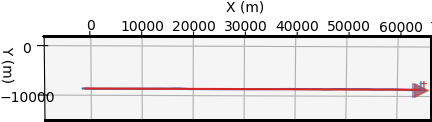}
\caption{Aircraft manoeuvring test scenario 1}
\end{figure}

\begin{figure}
\includegraphics[width=2in]{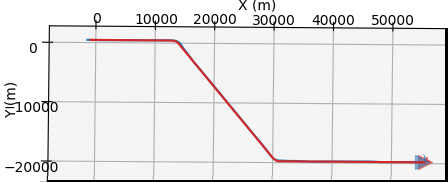}
\caption{Aircraft manoeuvring test scenario 2}
\end{figure}

\begin{figure}
\includegraphics[width=2in]{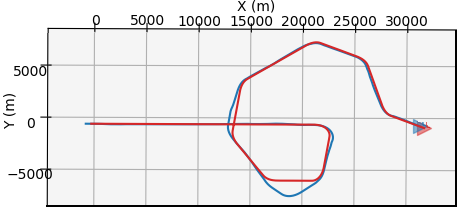}
\caption{Aircraft manoeuvring test scenario 3}
\end{figure}

\begin{figure}
\includegraphics[width=2in,height=1.3in]{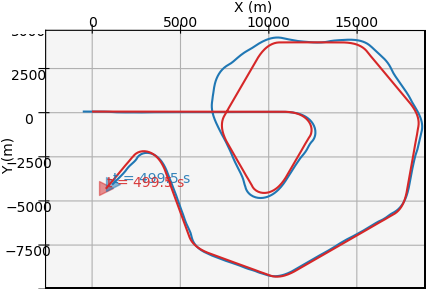}
\caption{Aircraft manoeuvring test scenario 4}
\end{figure}

Each episode is terminated after 700 timesteps and Table 3 shows the learning times for all the solutions. All values are relative to the average of Discrete 20,000. Discrete 40,000 takes about twice the time taken by Discrete 20,000, DQN 40,000 runs in about twice the time taken by DQN 20,000, and DDQN 40,000 completes in twice the time taken by DDQN 20,000. A3C 200,000 runs ten times longer than A3C 20,000. Because A3C uses four concurrent agents, learning takes much faster than its DQN and DDQN rivals. Among all the solutions being compared, D2D-SPL learns the fastest as it only takes 0.01\% longer than Discrete 20,000.

\begin{table*}
\centering
\begin{tabular}{|c|r|r|r|r|r|r|r|r|r|} 
\hline
Trial & Discrete & Discrete & \multicolumn{1}{|c|}{D2D} & \multicolumn{1}{|c|}{DQN} & \multicolumn{1}{|c|}{DQN} & \multicolumn{1}{|c|}{DDQN} & \multicolumn{1}{|c|}{DDQN} & \multicolumn{1}{|c|}{A3C} & \multicolumn{1}{|c|}{A3C}\\
      & \multicolumn{1}{|c|}{20,000} & \multicolumn{1}{|c|}{40,000} 
      & \multicolumn{1}{|c|}{SPL} 
      & \multicolumn{1}{|c|}{20,000} & \multicolumn{1}{|c|}{40,000} 
      & \multicolumn{1}{|c|}{20,000} & \multicolumn{1}{|c|}{40,000} 
      & \multicolumn{1}{|c|}{20,000} & \multicolumn{1}{|c|}{200,000}\\
\hline
0 & 0.9515 & 1.9079 & \textbf{0.9515} & 4.6971 & 11.1307 & 3.5559 & 7.2530 & 1.3936 & 13.5347\\
1 & 0.9476 & 1.8884 & \textbf{0.9476} & 4.8871 & 12.2939 & 3.4940 & 7.2707 & 1.3605 & 13.5817\\
2 & 1.1920 & 2.3880 & \textbf{1.1923} & 4.4365 & 13.8925 & 3.5478 & 7.2975 & 1.2748 & 13.4719\\
3 & 1.1778 & 2.3617 & \textbf{1.1779} & 4.5109 & 12.3102 & 4.4111 & 8.7122 & 1.2184 & 14.3311\\
4 & 1.2071 & 2.4377 & \textbf{1.2072} & 3.5728 & 7.3820 & 4.2812 & 8.3312 & 1.2382 & 14.0830\\
5 & 0.9106 & 1.8422 & \textbf{0.9106} & 3.5866 & 7.3969 & 4.1718 & 8.2209 & 1.3658 & 14.5852\\
6 & 0.9104 & 1.8391 & \textbf{0.9104} & 3.5898 & 7.3570 & 3.8600 & 7.7078 & 1.4976 & 13.8962\\
7 & 0.9127 & 1.8497 & \textbf{0.9127} & 3.5889 & 7.3631 & 3.7903 & 7.6311 & 1.5488 & 13.9328\\
8 & 0.8952 & 1.8074 & \textbf{0.8952} & 3.9266 & 8.9514 & 3.8597 & 7.8547 & 1.3989 & 14.9915\\
9 & 0.8952 & 1.8004 & \textbf{0.8952} & 3.8923 & 8.6911 & 3.8398 & 7.9101 & 1.3100 & 13.7658\\
\hline
Mean & 1.0000 & 2.0123 & \textbf{1.0001} & 4.0688 & 9.6769 & 3.8812 & 7.8189 & 1.3607 & 14.0174\\
\hline
\end{tabular}
\caption{Aircraft manoeuvring learning times relative to the mean of Discrete 20,000}
\end{table*}

Figure 7 shows the average reward per episode for all the solutions. The graphs have been smoothed-out by replacing every 200 consecutive rewards with their mean. It shows that using data from Discrete 20,000 to train the D2D-SPL network results in an average score of 0.95 when the training set is re-applied to the resulting network. This score is much higher than the scores of the other methods.

It can also seen that DQN suffers from overestimation as reported in \cite{vanhasselt2016}, which adversely affects the policy. Overestimation did not occur in DDQN.

\begin{figure*}
\includegraphics[width=6in,height=1.4in]{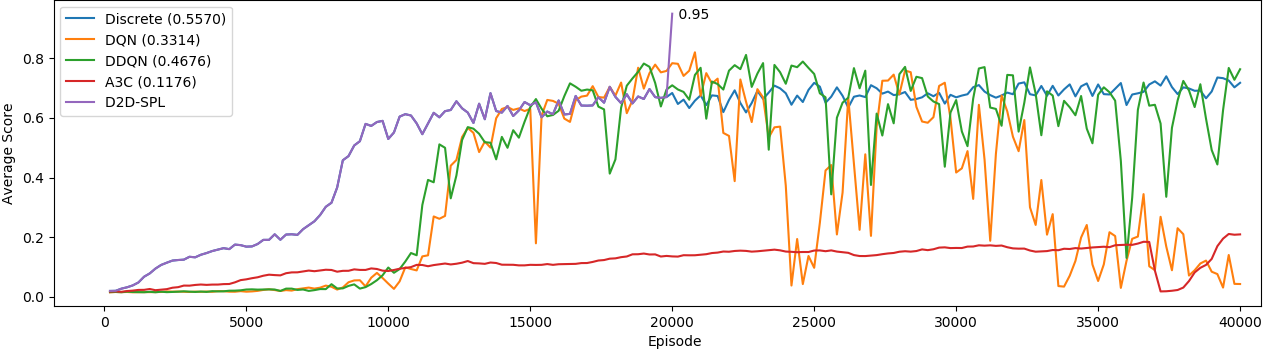}
\caption{Reward/episode for aircraft manoeuvring for all solutions (200:1)}
\end{figure*}

Tables 4 to 7 show the results for the four test scenarios. In the discrete cases, Discrete 40,000 is generally better than Discrete 20,000 because the learning curve with actor-critic is more stable, not as noisy as the DQN or A3C. This means, longer learning tends to produce a better policy. In all cases, D2D-SPL performs better than its base Discrete 20,000 and even Discrete 40,000. It is also better than DQN 20,000 in three of four cases and better than DQN 40,000 in all cases. Because of the possible overestimation in DQN, there is no guarantee that longer learning (in this case, DQN 40,000) will produce a better model than shorter learning (in this case, DQN 20,000).

\begin{table*}
\centering
\begin{tabular}{|c|r|r|r|r|r|r|r|r|r|} 
\hline
Trial & Discrete & Discrete & \multicolumn{1}{|c|}{D2D} & \multicolumn{1}{|c|}{DQN} & \multicolumn{1}{|c|}{DQN} & \multicolumn{1}{|c|}{DDQN} & \multicolumn{1}{|c|}{DDQN} & \multicolumn{1}{|c|}{A3C} & \multicolumn{1}{|c|}{A3C}\\
      & \multicolumn{1}{|c|}{20,000} & \multicolumn{1}{|c|}{40,000} 
      & \multicolumn{1}{|c|}{SPL} 
      & \multicolumn{1}{|c|}{20,000} & \multicolumn{1}{|c|}{40,000} 
      & \multicolumn{1}{|c|}{20,000} & \multicolumn{1}{|c|}{40,000} 
      & \multicolumn{1}{|c|}{20,000} & \multicolumn{1}{|c|}{200,000}\\
\hline
0 & 0.9123 & 0.9182 & \textbf{0.9502} & 0.9065 & 0.0259 & 0.7816 & 0.5363 & 0.0478 & 0.2515\\
1 & 0.8142 & 0.8142 & \textbf{0.8956} & 0.8748 & 0.4256 & 0.0558 & 0.1104 & 0.1430 & 0.0485\\
2 & 0.8508 & \textbf{0.9130} & 0.7656 & 0.2960 & 0.0327 & 0.0755 & 0.4394 & 0.1519 & 0.1575\\
3 & 0.9274 & 0.9236 & \textbf{0.9312} & 0.0250 & 0.0327 & 0.8304 & 0.0603 & 0.1611 & 0.0478\\
4 & 0.9086 & 0.9205 & \textbf{0.9578} & 0.8620 & 0.9627 & 0.8325 & 0.6189 & 0.1638 & 0.0813\\
5 & \textbf{0.9899} & \textbf{0.9899} & 0.9572 & 0.7936 & 0.0286 & 0.0453 & 0.8748 & 0.1681 & 0.8710\\
6 & 0.8142 & 0.8142 & 0.9490 & \textbf{0.9687} & 0.0256 & 0.9174 & 0.1000 & 0.1599 & 0.8748\\
7 & 0.9237 & 0.9204 & \textbf{0.9530} & 0.7928 & 0.1620 & 0.6233 & 0.5615 & 0.1689 & 0.1026\\
8 & 0.8142 & 0.8142 & \textbf{0.9007} & 0.0412 & 0.1170 & 0.8995 & 0.0507 & 0.2033 & 0.2055\\
9 & 0.8142 & 0.8142 & \textbf{0.9463} & 0.9211 & 0.8751 & 0.8748 & 0.7799 & 0.0478 & 0.1298\\
\hline
Mean & 0.8769 & 0.8842 & \textbf{0.9207} & 0.6482 & 0.2688 & 0.5936 & 0.4132 & 0.1416 & 0.2770\\
\hline
Median & 0.8797 & 0.9156 & \textbf{0.9476} & 0.8278 & 0.0748 & 0.8060 & 0.4879 & 0.1605 & 0.1437\\
\hline
\end{tabular}
\caption{Aircraft manoeuvring test results for test scenario 1}
\end{table*}

\begin{table*}
\centering
\begin{tabular}{|c|r|r|r|r|r|r|r|r|r|} 
\hline
Trial & Discrete & Discrete & \multicolumn{1}{|c|}{D2D} & \multicolumn{1}{|c|}{DQN} & \multicolumn{1}{|c|}{DQN} & \multicolumn{1}{|c|}{DDQN} & \multicolumn{1}{|c|}{DDQN} & \multicolumn{1}{|c|}{A3C} & \multicolumn{1}{|c|}{A3C}\\
      & \multicolumn{1}{|c|}{20,000} & \multicolumn{1}{|c|}{40,000} 
      & \multicolumn{1}{|c|}{SPL} 
      & \multicolumn{1}{|c|}{20,000} & \multicolumn{1}{|c|}{40,000} 
      & \multicolumn{1}{|c|}{20,000} & \multicolumn{1}{|c|}{40,000} 
      & \multicolumn{1}{|c|}{20,000} & \multicolumn{1}{|c|}{200,000}\\
\hline
0 & 0.9107 & 0.6160 & \textbf{0.9379} & 0.8564 & 0.0259 & 0.7838 & 0.3568 & 0.0478 & 0.2372\\
1 & 0.7999 & 0.7999 & \textbf{0.8923} & 0.3352 & 0.3411 & 0.0500 & 0.0567 & 0.1689 & 0.0478\\
2 & 0.7007 & \textbf{0.8771} & 0.8059 & 0.0806 & 0.0327 & 0.0762 & 0.4437 & 0.1623 & 0.1306\\
3 & 0.8747 & 0.8329 & \textbf{0.9208} & 0.0250 & 0.0327 & 0.8716 & 0.0603 & 0.1779 & 0.0478\\
4 & 0.6414 & 0.8424 & 0.9470 & 0.8459 & \textbf{0.9580} & 0.8449 & 0.4214 & 0.1618 & 0.0800\\
5 & \textbf{0.9054} & 0.8895 & 0.7539 & 0.7894 & 0.0286 & 0.0397 & 0.2087 & 0.1418 & 0.8560\\
6 & 0.7999 & 0.7999 & 0.8413 & \textbf{0.9376} & 0.0256 & 0.9096 & 0.1002 & 0.2059 & 0.2092\\
7 & \textbf{0.8858} & 0.8662 & 0.7887 & 0.7832 & 0.1571 & 0.1772 & 0.5615 & 0.1616 & 0.1333\\
8 & 0.7999 & 0.7999 & \textbf{0.8183} & 0.0412 & 0.1272 & 0.6397 & 0.0522 & 0.2410 & 0.1427\\
9 & \textbf{0.8559} & 0.8458 & 0.8515 & 0.3516 & 0.8442 & 0.8529 & 0.4823 & 0.0478 & 0.1919\\
\hline
Mean & 0.8174 & 0.8170 & \textbf{0.8558} & 0.5046 & 0.2573 & 0.5246 & 0.2744 & 0.1517 & 0.2077\\
\hline
Median & 0.8279 & 0.8377 & \textbf{0.8464} & 0.5674 & 0.0799 & 0.7117 & 0.2828 & 0.1620 & 0.1380\\
\hline
\end{tabular}
\caption{Aircraft manoeuvring test results for test scenario 2}

\vspace{-0.15cm}

\centering
\begin{tabular}{|c|r|r|r|r|r|r|r|r|r|} 
\hline
Trial & Discrete & Discrete & \multicolumn{1}{|c|}{D2D} & \multicolumn{1}{|c|}{DQN} & \multicolumn{1}{|c|}{DQN} & \multicolumn{1}{|c|}{DDQN} & \multicolumn{1}{|c|}{DDQN} & \multicolumn{1}{|c|}{A3C} & \multicolumn{1}{|c|}{A3C}\\
      & \multicolumn{1}{|c|}{20,000} & \multicolumn{1}{|c|}{40,000} 
      & \multicolumn{1}{|c|}{SPL} 
      & \multicolumn{1}{|c|}{20,000} & \multicolumn{1}{|c|}{40,000} 
      & \multicolumn{1}{|c|}{20,000} & \multicolumn{1}{|c|}{40,000} 
      & \multicolumn{1}{|c|}{20,000} & \multicolumn{1}{|c|}{200,000}\\
\hline
0 & 0.3760 & 0.3463 & 0.8334 & \textbf{0.8378} & 0.0259 & 0.6751 & 0.2139 & 0.0478 & 0.2988\\
1 & 0.4326 & \textbf{0.7685} & 0.7123 & 0.4427 & 0.1853 & 0.0558 & 0.1496 & 0.1755 & 0.0478\\
2 & 0.7350 & \textbf{0.7476} & 0.2928 & 0.2056 & 0.0327 & 0.0755 & 0.5904 & 0.1607 & 0.0556\\
3 & \textbf{0.7490} & 0.6128 & 0.7393 & 0.0250 & 0.0327 & 0.8472 & 0.0623 & 0.1756 & 0.0478\\
4 & 0.4257 & 0.4731 & 0.8563 & 0.8399 & \textbf{0.9223} & 0.8179 & 0.2360 & 0.1806 & 0.0816\\
5 & 0.4436 & 0.4917 & 0.3530 & \textbf{0.7523} & 0.0286 & 0.0474 & 0.3116 & 0.1585 & 0.6567\\
6 & 0.6452 & 0.3009 & 0.4664 & \textbf{0.8984} & 0.0256 & 0.8827 & 0.1000 & 0.1778 & 0.3118\\
7 & 0.3349 & 0.6714 & 0.6665 & \textbf{0.7423} & 0.1576 & 0.2888 & 0.4888 & 0.1702 & 0.1271\\
8 & 0.3562 & \textbf{0.3962} & 0.3290 & 0.0412 & 0.0852 & 0.7296 & 0.0508 & 0.2594 & 0.1800\\
9 & 0.3893 & 0.4662 & 0.3332 & 0.3484 & 0.7212 & \textbf{0.8453} & 0.3092 & 0.0478 & 0.1912\\
\hline
Mean & 0.4887 & 0.5275 & \textbf{0.5582} & 0.5133 & 0.2217 & 0.5265 & 0.2512 & 0.1554 & 0.1998\\
\hline
Median & 0.4291 & 0.4824 & 0.5664 & 0.5925 & 0.0589 & \textbf{0.7024} & 0.2249 & 0.1729 & 0.1535\\
\hline
\end{tabular}
\caption{Aircraft manoeuvring test results for test scenario 3}

\vspace{-0.15cm}

\centering
\begin{tabular}{|c|r|r|r|r|r|r|r|r|r|} 
\hline
Trial & Discrete & Discrete & \multicolumn{1}{|c|}{D2D} & \multicolumn{1}{|c|}{DQN} & \multicolumn{1}{|c|}{DQN} & \multicolumn{1}{|c|}{DDQN} & \multicolumn{1}{|c|}{DDQN} & \multicolumn{1}{|c|}{A3C} & \multicolumn{1}{|c|}{A3C}\\
      & \multicolumn{1}{|c|}{20,000} & \multicolumn{1}{|c|}{40,000} 
      & \multicolumn{1}{|c|}{SPL} 
      & \multicolumn{1}{|c|}{20,000} & \multicolumn{1}{|c|}{40,000} 
      & \multicolumn{1}{|c|}{20,000} & \multicolumn{1}{|c|}{40,000} 
      & \multicolumn{1}{|c|}{20,000} & \multicolumn{1}{|c|}{200,000}\\
\hline
0 & 0.5641 & 0.3044 & \textbf{0.8093} & 0.8006 & 0.0283 & 0.2382 & 0.1329 & 0.0478 & 0.2773\\
1 & 0.2615 & 0.3348 & \textbf{0.3915} & 0.2863 & 0.1389 & 0.0501 & 0.0506 & 0.1758 & 0.0478\\
2 & 0.1841 & 0.4393 & 0.1782 & 0.0774 & 0.0327 & 0.0949 & \textbf{0.5106} & 0.1938 & 0.1976\\
3 & 0.1992 & 0.2574 & 0.1892 & 0.0250 & 0.0327 & \textbf{0.8437} & 0.0609 & 0.1844 & 0.0478\\
4 & 0.1859 & 0.1948 & \textbf{0.8340} & 0.8197 & 0.6820 & 0.7781 & 0.3408 & 0.2020 & 0.0789\\
5 & 0.1987 & 0.3113 & 0.1994 & \textbf{0.7693} & 0.0371 & 0.0414 & 0.1715 & 0.1384 & 0.6853\\
6 & 0.2172 & 0.5787 & 0.2083 & 0.8142 & 0.0256 & \textbf{0.8565} & 0.1019 & 0.1915 & 0.1717\\
7 & 0.1933 & 0.2045 & 0.7529 & \textbf{0.8031} & 0.1510 & 0.1750 & 0.5824 & 0.1637 & 0.1226\\
8 & 0.1744 & 0.6813 & 0.1978 & 0.0412 & 0.0858 & \textbf{0.8391} & 0.0627 & 0.2881 & 0.2561\\
9 & 0.1808 & 0.2318 & 0.2194 & 0.2001 & 0.5020 & \textbf{0.8385} & 0.1809 & 0.0478 & 0.1654\\
\hline
Mean & 0.2359 & 0.3538 & 0.3980 & 0.4637 & 0.1716 & \textbf{0.4756} & 0.2195 & 0.1633 & 0.2050\\
\hline
Median & 0.1960 & 0.3079 & 0.2138 & \textbf{0.5278} & 0.0615 & 0.5082 & 0.1522 & 0.1801 & 0.1685\\
\hline
\end{tabular}
\caption{Aircraft manoeuvring test results for test scenario 4}
\end{table*}

\section{Conclusions and Future Work}
We build and test solutions for Cartpole and an aircraft manoeuvring simulator. The difference among the solutions is the learning time and the performance of the generated policy or model. It is shown that actor-critic works for both problems and longer learning with more episodes tends to produce a better policy, even though, as shown in Figure 7, the result is unstable in the sense that the policy after $n$ episodes is not always better than the policy before that. This means, when we decide to stop learning after episode $n$, we need to record $m$ policies before episode $n$, test them against some pre-set criteria, and choose the best policy. For instance, if we decide to run a learning session for 20,000 episodes, we might want to compare all policies resulting from the last 500 episodes or so.

We also show that D2D-SPL can shorten the learning time of the actor-critic algorithm. The D2D-SPL results are consistently better than the base policy used to train it and even better than the policy obtained by resuming the actor-critic learning. In addition, because D2D-SPL gets its data from the top 5\% of the episodes, the resulting model is stable. As D2D-SPL uses a neural-network, which generally is known to be a good function approximator, it is not surprising that D2D-SPL performs better than discrete actor-critic in generalising test cases not seen during learning.

In both Cartpole and Aircraft Manoeuvring, DQN, DDQN and A3C can be used to train a neural network. However, D2D-SPL learns much faster and performs better in Cartpole and in the majority of the test cases in Aircraft Manoeuvring than its competitors.

One difficulty in using D2D-SPL is to find a discretisation scheme that leads to a good policy. In the case of Cartpole, the discretisation scheme has been made available in \cite{Barto1983}. In the case of Aircraft Manoeuvring it took us many experiments to come up with a good discretisation scheme. Generally, the more state variables there are, the more combinations there are that are possible and the harder to get it right. Future studies may focus on using D2D-SPL in environments with higher numbers of state variables.

Since this is the first time D2D-SPL is ever used, there are a number of areas of where further work with D2D-SPL can be undertaken. Future work may utilise other tabular RL algorithms, such as Q-learning and SARSA in the RL part of the system. It is also possible to further train the resulting network of D2D-SPL, using an existing or new method, to improve performance. Extending this work to apply to more complex domains which are inherently multi-objective is also of interest \cite{Roijers2013} \cite{Vamplew2011}.

\begin{acks}
This research is supported by the Defence Science and Technology Group, Australia; the Defence Science Institute, Australia; and an Australian Government Research Training Program Fee-offset scholarship. Associate Professor Joarder Kamruzzaman of the Centre for Multimedia Computing, Communications, and Artificial Intelligence Research (MCCAIR) at Federation University contributed some of the computing resources for this project.
\end{acks}
\clearpage
\bibliographystyle{ACM-Reference-Format}  
\bibliography{references}  

\end{document}